%% file: acl_latex.tex
\pdfoutput=1

\documentclass[11pt]{article}

\usepackage[preprint]{acl}

\usepackage{times}
\usepackage{latexsym}

\usepackage[T1]{fontenc}

\usepackage[utf8]{inputenc}

\usepackage{microtype}

\usepackage{inconsolata}

\usepackage{graphicx}

\usepackage{cleveref}
\usepackage[inline]{enumitem}
\usepackage{caption}
\usepackage{subcaption}
\usepackage{booktabs}
\usepackage{tcolorbox}
\usepackage{siunitx}
\usepackage{multirow}
\usepackage{rotating}
\usepackage{makecell}
\usepackage{adjustbox}
\usepackage{tipa} 

\newcommand\cbox[2]{\noindent\begin{tcolorbox}[width=\linewidth,colback={lightgray!40},size=small]\scalebox{-1}[1]{\includegraphics[height=1.1\fontcharht\font`\B]{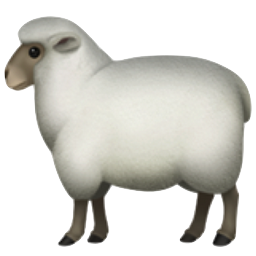}} \textbf{#1}: #2\end{tcolorbox}}

\newcommand\uponacceptance[1]{\textcolor{gray!40}{#1}}

\title{LLäMmlein \includegraphics[height=1.2\fontcharht\font`\B]{pics/sheep.png}: Transparent, Compact and Competitive\\German-Only Language Models from Scratch}

\author{Jan Pfister$^{\includegraphics[height=1.2\fontcharht\font`\B]{pics/sheep.png}}$ \and Julia Wunderle$^{\includegraphics[height=1.2\fontcharht\font`\B]{pics/sheep.png}}$ \and Andreas Hotho \\
Data Science Chair\\
Center for Artificial Intelligence and Data Science (CAIDAS)\\
Julius-Maximilians-Universität Würzburg (JMU)\\
\texttt{\{lastname\}@informatik.uni-wuerzburg.de}\\}

\begin{document}
\maketitle
\makeatletter
\def\blfootnote{\gdef\@thefnmark{}\@footnotetext}
\makeatother
\blfootnote{\includegraphics[height=2\fontcharht\font`\B]{pics/sheep.png} These authors contributed equally to this work.}

\begin{abstract}
        We transparently create two German-only decoder models, LLäMmlein 120M and 1B\footnote{After submission we also released a new 7B model}, from scratch and publish them, along with training data, for the (German) NLP research community to use\footnote{
                \url{https://professor-x.de/lm/LLaMmlein}
        }.
        The model training involved several key steps, including data preprocessing/filtering, the creation of a German tokenizer, the training itself, as well as the evaluation of the final models on various benchmarks, also against existing models.
        Throughout the training process, multiple checkpoints were saved in equal intervals and analyzed using the German SuperGLEBer benchmark to gain insights into the models' learning process.

        Compared to state-of-the-art models on the SuperGLEBer benchmark, both LLäMmlein models performed competitively, consistently matching or surpassing models with similar parameter sizes.
        The results show that the models' quality scales with size as expected, but performance improvements on some tasks plateaued early during training, offering valuable insights into resource allocation for future models.
\end{abstract}

\section{Introduction}
Large Language Models (LLMs) have achieved remarkable success, yet this progress is predominantly centered on English.
Other languages, including German, lag behind due to limited competition, reduced investment, and a lack of transparency in training data, code, and detailed results (also see \citealp{pfister-hotho-2024-supergleber}).
While smaller German-only models do exist, such as BERTs or smaller GPTs \cite{chan-etal-2020-germans,Scheible2020GottBERTAP}, or the contemporaneous effort by \citet{dosmo7b2024}, many are closed or undocumented, and few robust, openly accessible German LLMs have been built from scratch with full transparency.
Most German-capable LLMs are either multilingual models (e.g., mGPT~\cite{mGPT}) or English models adapted to German (e.g., LeoLM~\cite{pluster2023leolm}, BübleLM~\cite{delobelle2024buble}, or bloom-clp~\cite{Ostendorff2023clp}).
Others, like Mistral~\cite{jiang2023mistral7b}, share little about their (German) training data, making it difficult to understand which resources and pretraining strategies are most suitable for developing strong German LLMs from scratch.
This lack of traceability hampers the community's ability to identify, curate, and refine German data, and to examine how corpus and training choices affect model quality.
Subtle issues -- such as performance deteriorations on non-English downstream tasks \cite{virtanen2019multilingual} or poor handling of German's complex grammar and morphology (\citet{mielke-etal-2019-kind}, also \Cref{app:exemp_shortcomings}) -- remain common, even in state-of-the-art models like Llama 3~\cite{dubey2024llama3herdmodels}, which can revert to English despite a German context, or sometimes sound like machine-translated from English\footnote{\url{https://www.reddit.com/r/LocalLLaMA/comments/1bfce18/still_didnt_found_a_better_small_german_llm_anyone/}}.

We present \emph{LLäMmlein}, the first German-only LLM family trained entirely from scratch with full transparency, providing a foundation for systematically analyzing the relationship between training data and model outputs.
To this end, we share the model, code, and dataset to foster reproducibility and collaboration.
Although our evaluation is primarily illustrative, it offers insights into the model's German capabilities and enables further research and development.
We accompany training with iterative benchmark evaluations, tracking the learning progress of our 120M and 1B model to illuminate scaling effects and guide future research.

To achieve this, we:
\begin{enumerate*}[label=(\arabic*)]
        \item clean and filter a large German dataset derived from RedPajama V2 \cite{together2023redpajama}, ensuring high-quality input,
        \item construct a dedicated German tokenizer (32k tokens) fitted on varying data amounts to compare against existing German tokenizers,
        \item pretrain two exclusively German autoregressive LLMs (120M and 1B) and release incremental checkpoints, inspired by \citet{pythia}, to inform efficient stopping criteria and shed light on learning dynamics, and
        \item evaluate the models on a range of tasks (SuperGLEBer \cite{pfister-hotho-2024-supergleber}, lm-evaluation-harness-de \cite{leo_translate,eval-harness}) to benchmark performance against existing models.
\end{enumerate*}

In doing so, we directly demonstrate and address the pressing need for dedicated German-centric LLM research, establishing a transparent foundation for understanding, improving, and expanding the German LLM ecosystem.
\cbox{}{Throughout the paper, we highlight interesting findings and insights we gained during the process in little boxes like this one.}

\section{Methodology}
Pretraining and evaluating a German LLM from scratch, end-to-end involves several steps, including dataset preprocessing (\cref{sec:dataset_preproc}), tokenizer fitting (\cref{sec:tokenizer}), model pretraining (\cref{sec:pretraining}), model evaluation using a comprehensive German benchmark, as well as multiple translated prompt-based few-shot QA tasks (\cref{sec:meth_evaluation}), and exemplary downstream adaptations (\cref{sec:downstream}).

\subsection{Dataset}\label{sec:dataset}
RedPajama V2 is an open\footnote{\url{https://commoncrawl.org/terms-of-use}}, multilingual dataset designed for training large language models \cite{together2023redpajama}.
It consists of over 100 billion text documents collected from 84 CommonCrawl snapshots between 2014 and 2023 and encompasses multiple languages, including English, German, French, Italian, and Spanish.
The dataset was originally preprocessed using the CCNet pipeline \cite{wenzek-etal-2020-ccnet} leading to about 30 billion overall documents further enriched with quality signals and duplicate indicators.
Using perplexity of a language model, the RedPajama dataset was divided into three quality categories, in descending order of quality: head, middle, and tail.
Following a manual inspection of a randomly selected subset, the head and middle partitions were deemed to contain sufficiently high-quality German texts suitable for continued use.
In contrast, the tail partition exhibited inconsistent quality and was consequently excluded from further training.

\subsubsection{Dataset Analysis}\label{sec:dataset_analysis}
The aim of the following preliminary dataset analysis is to gain a deeper understanding of the German portion of the dataset used.
The ``official'' estimate of the size of the German segment within the RedPajamaV2 dataset, derived through extrapolation from a smaller sample analyzed with Mistral-7B, is approximately 3 trillion German tokens~\cite{together2023redpajama}.
Following, we first perform an exploratory analysis of the dataset to gain a clearer understanding of the actual amount of German data it contains, alongside its domain distribution and the most prevalent data sources.

\begin{figure}
        \centering
        \includegraphics[width=\linewidth]{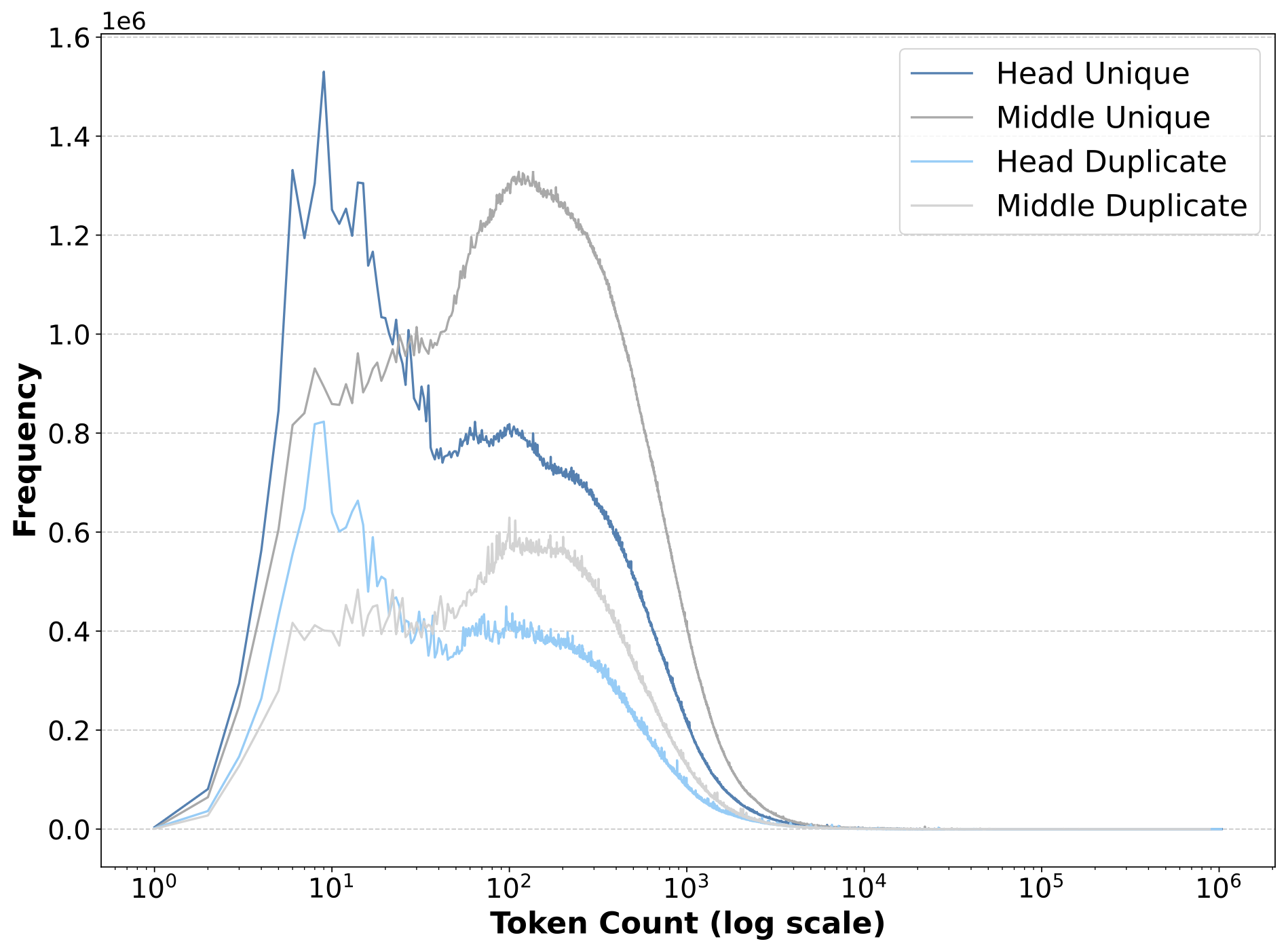}
        \caption{Token count distribution for each partition separately: head unique, middle unique, head duplicate and middle duplicate based on gbert-large tokenizer}\label{fig:redpajama_unique}
\end{figure}

\paragraph{Statistics} Our own count using the gbert-large tokenizer, led to a token count of 2.7 trillion German tokens for head and middle combined, before document-level deduplication.

\Cref{fig:redpajama_unique} breaks down the distribution of each partition, i.e.\ head unique, middle unique, head duplicate and middle duplicate separately.
The first occurence of a document is considered unique, while all subsequent appearances are marked as duplicates.
The middle unique partition contains the largest amount of data, with approximately 1.2 billion samples, which corresponds to 45\% of the full dataset.
The head unique partition, by comparison, includes around 400 million fewer samples.

Overall, most samples are unique (1.9 billion samples) and only significantly less are marked as duplicates, appearing a second or more times (777 million samples) across the entire dataset.
The most common token-per-document count can be found at nine, with approximately 3.6 million occurrences in the dataset.
A second peak (most prominent for the unique split) occurs at around 100 tokens per document.
In total, the 2.7 trillion German tokens are distributed across samples with lengths ranging from 1 to \num{1034799} tokens, averaging approximately \num{1000} tokens per sample.

\begin{figure}
        \centering
        \includegraphics[width=\linewidth]{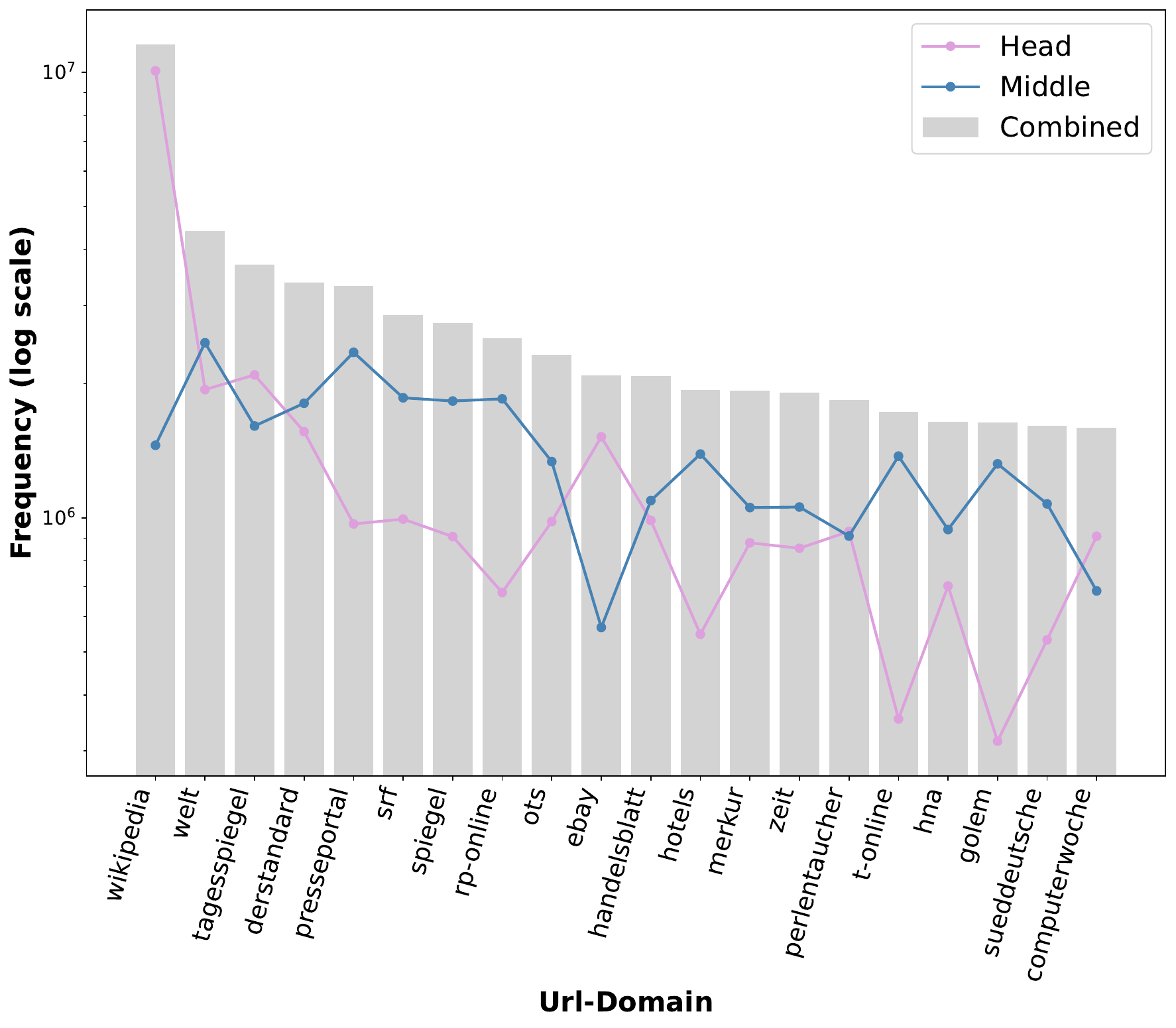}
        \caption{Top 20 most frequent domains across the full dataset in gray with frequencies in head and middle partitions separately.}\label{fig:domains}
\end{figure}

\paragraph{Domain Analysis} The dataset contains content crawled from various domain names.
\Cref{fig:domains} displays the top 20 sources from which the data was collected, with the overall count illustrated as gray bars and separate plots for the head (pink) and middle (blue) unique splits.

Wikipedia clearly stands out as the largest contributor, with a combined total of over 11.5 million samples.
Among these, about 10 million entries belong to the head category, while about 1.45 million stem from the middle partition.
This distribution aligns with the fact that the split into head, middle and tail was created using a perplexity score criterion based on a language model trained on Wikipedia~\cite{together2023redpajama} -- consequently, texts closer in style to Wikipedia tend to be ranked higher.
Besides Wikipedia, it is evident that news websites also constitute a significant portion of the dataset.
For the middle split, ``welt.de'' emerges as the most frequent domain, contributing around 2.47 million samples.
With the exception of domains like eBay, hotels and Perlentaucher, the list is largely dominated by general news outlets.

\subsubsection{Further Dataset Preprocessing}\label{sec:dataset_preproc}
To remove common web boilerplate, such as EU-specific ``General Data Protection Regulation'' (GDPR) notices or similar repetitive content, we utilize a paragraph-level deduplication scheme powered by Dolma -- a framework that enables efficient deduplication through a Rust-based Bloom filter \cite{soldaini-etal-2024-dolma}.
A Bloom filter is a probabilistic method that works similarly to a hash table, determining whether an element has already been previously encountered or not.
This ensures that highly redundant text is filtered out, improving the overall quality and diversity of the dataset.
This approach may inadvertently over-remove valid and relevant content, such as short texts mistakenly treated as entire paragraphs being removed across the dataset.
To mitigate this, and to preserve meaningful short text sections -- such as lists or frequently occurring itemized phrases that are contextually significant -- we excluded paragraphs containing fewer than three words from the deduplication process.

Despite these efforts, we found that some unusual artifacts, such as long sequences of guitar chords, remained, as they scored low perplexity (41.2) compared to the average perplexity of 206.35 in the respective snapshot (2014\_52) and were therefore not removed by the preliminary quality filter.
To address this, we built a token-to-word ratio filter.
Here, we compared the word count (whitespace separated) with the token count using the German GPT-2 tokenizer \cite{german-gpt2}.
According to our intuition, a usual ratio between the two counts indicates abnormal or low-quality text, whereas a close match suggests valid German content.
A simple example illustrates this clearly:
The phrase ``Der Himmel ist blau'' consists of 4 words and 4 tokens, so it is not removed by our filter.
In contrast, ``/de/c/trebic-unesco'' counts as 1 word but 11 tokens, and should therefore be excluded by this token-to-word ratio filter.

Preliminary examinations and manual review suggested a ratio of tokens to words of eight as a valid threshold.
Thus, paragraphs exceeding this threshold were excluded from the dataset.
\cbox{Interesting}{Regular patterns of non-textual data, such as guitar chords, can yield a low perplexity and therefore remain in the dataset through quality filtering processes.}

\subsection{Model Pretraining Framework}\label{sec:pretraining}
While there are several existing resources and repositories for training an LLM from scratch\footnote{a small subset: \url{https://github.com/Hannibal046/Awesome-LLM\#llm-training-frameworks}}, we chose the TinyLlama GitHub repository as the backbone of our project~\cite{zhang2024tinyllama}.
It was used to pretrain the 1B English Llama 2-based~\cite{touvron2023llama2openfoundation} TinyLlama model from scratch before and builds upon the lit-gpt repository \cite{lit-gpt}, which provides robust tooling for data preparation, fine-tuning, pretraining, and deploying LLMs using PyTorch Lightning.

It includes all key features such as multi-GPU and multi-node distributed training with FSDP as well as FlashAttention-2~\cite{dao2023flashattention2}.
In addition, it provides scripts to convert the models into HuggingFace format for easier use and distribution.

We modified the codebase\footnote{\url{https://github.com/LSX-UniWue/LLaMmlein}} for our requirements:
\begin{enumerate*}
        \item We significantly improved the data loader speed by adding various layers of caching.
        \item We enable training directly from a directory of jsonl-files, without any prior preprocessing.
        \item Most importantly, inspired by our reviewers' feedback, we retrained our models logging exact datapoints as they enter the model.
              This allows us to correlate the training data and its order for each of our published (intermediate) checkpoints\footnote{Since the performance of the original model (without data logging) and the new model (with data logging) did not differ significantly, we kept the original scores throughout the paper.}.
\end{enumerate*}

\subsection{Model Evaluation Setup}\label{sec:meth_evaluation}
\subsubsection{Intermediate Checkpoint Evaluation} \label{sec:checkpoint_eval}
To get a better understanding of the training, we monitor the progress and regularly evaluate intermediate checkpoints on six representative SuperGLEBer tasks \cite{pfister-hotho-2024-supergleber} following a finetuning to the task from the respective checkpoint.
These tasks were selected to encompass a range of problem types to assess our model's performance.
Within classification,
\begin{enumerate*}[label=(\arabic*)]
        \item Natural Language Inference(NLI) \cite{conneau-etal-2018-xnli} requires determining whether a hypothesis is entailed, neutral, or contradictory to a premise;
        \item FactClaiming Comments \cite{germeval-2021-germeval} involves binary classification of fact-checkable claims;
        \item DB Aspect \cite{Wojatzki2017GermEval2S} addresses multi-label categorization and polarity detection in input sentences; and
        \item WebCAGe \cite{henrich-etal-2012-webcage} tests if a given word's sense aligns across two contexts.
              For sequence tagging,
        \item EuroParl \cite{Faruqui2010TrainingAE} evaluates Named Entity Recognition on European Parliament data.
              Finally, in the sentence similarity domain,
        \item PAWSX \cite{liang-etal-2020-xglue} challenges the model to detect paraphrases via vector representations.

\end{enumerate*}

\subsubsection{Final Model Evaluation}
To assess general knowledge and abilities, we evaluated our final models on the full SuperGLEBer benchmark (29 tasks across classification, sequence tagging, question answering, and sentence similarity) \cite{pfister-hotho-2024-supergleber}, as well as machine-translated, prompt-based, few-shot QA tasks using the lm-evaluation-harness-de by \citet{leo_translate} (if not stated otherwise, they are evaluated measuring unnormalized and Byte-length normalized accuracy \cite{eval-harness}):
\begin{enumerate*}[label=(\arabic*)]\label{sec:leo_eval}
        \item ARC-Challenge-DE \cite{Clark2018ThinkYH}: Grade-school science questions (\num{1471} samples).
              Few-shot evaluation with 25 samples.\label{sec:mmlu}
        \item MMLU-DE \cite{hendryckstest2021}: \num{6829} multiple-choice questions across 57 topics (e.g., math, medicine, law).
              Few-shot evaluation with five samples.
        \item HellaSwag-DE \cite{zellers2019hellaswag}: Commonsense reasoning dataset with \num{11000} translated samples, featuring incomplete sentences with multiple-choice completions.
              Few-shot evaluation with ten samples.
        \item TruthfulQA-DE \cite{lin-etal-2022-truthfulqa}: 817 questions across 38 categories designed to evaluate model truthfulness, particularly in handling misconceptions in a zero-shot evaluation setting.
              Performance here is measured using MC1 (Single-true): accuracy in selecting a single correct answer and MC2 (Multi-true): the normalized probability assigned to all correct answers \cite{eval-harness}.
\end{enumerate*}

To assess the impact of checkpoint averaging on model performance~\cite{vaswani_attention_2017,dubey2024llama3herdmodels}, we also evaluate averaged checkpoints on the SuperGLEBer benchmark.

\subsection{Exemplary Downstream Adaptations}\label{sec:downstream}
As examples of downstream adaptation, we finetune our model with LoRA~\cite{hu2021loralowrankadaptationlarge} in two settings:
instruct-tuning (adapting the model to respond to user prompts) and for demonstration purposes on Bavarian and Swiss in \Cref{app:dialectic_exploration}.

\section{Experiments}
\subsection{Tokenizer}\label{sec:tokenizer}

We follow the TinyLlama setup and fit an Llama 2 Byte-Pair Encoding (BPE) tokenizer with a \num{32000}-token vocabulary \cite{touvron2023llama2openfoundation}.
We trained three tokenizers on different amounts of data:

\begin{enumerate*}[label=(\arabic*)]
        \item 1TB: Spans backward from the most recent data until 1TB of data is processed
        \item 2023-2021: Includes all splits of the high quality data split from years 2023 to 2021 (847GB)
        \item 2023\_14: Consists of the most recent 2023\_14 split (67GB)
\end{enumerate*}

\subsection{Model Pretraining}
We trained two models, LLäMmlein 120M and LLäMmlein 1B, using filtered subsets of our preprocessed dataset (\cref{sec:dataset}).
Detailed configurations for both models are provided in \Cref{tab:architectures}, and \Cref{fig:loss_120M,fig:loss_1b} display the model's loss curve, where each training resume is distinguished by a unique color.

\subsubsection{LLäMmlein 120M}
LLäMmlein 120M was trained on the filtered head unique partition, comprising 1T pretraining tokens (2 epochs).
The training setup included a maximum learning rate of 6e-4, grouped query attention of 4, a sequence length of \num{2048}, and a global batch size of 1024.
We employed the full-shard FSDP strategy across 32 L40 GPUs on 16 nodes, completing training in approximately \num{10000} GPU hours.

\subsubsection{LLäMmlein 1B}

LLäMmlein 1B was trained on the filtered, head and middle unique partitions, resulting in a dataset of 1.3T unique tokens, and 3T overall tokens seen during training.
The training setup featured a maximum learning rate of 6e-4, a global batch size of 1024 (per device batch size of 16), and was executed on 64 A100 80GB GPUs across 8 nodes over 32 days (\num{50000} GPU hours).

\subsection{Downstream Adaptations}\label{sec:exp_downstream}
For instruct-tuning, we use LoRA~\cite{hu2021loralowrankadaptationlarge} with supervised finetuning, PEFT~\cite{peft} and default hyperparameters for three epochs on the following datasets from huggingface:
``LSX-UniWue/Guanako'', ``FreedomIntelligence/alpaca-gpt4-deutsch'', ``FreedomIntelligence/evol-instruct-deutsch'', and ``FreedomIntelligence/sharegpt-deutsch''.
We train and publish a separate adapter for each dataset, and also a combined model across all datasets.

\section{Evaluation}\label{sec:evaluation}
\subsection{Tokenizer}\label{sec:eval_tokenizer}
We evaluate our custom Llama2-based tokenizers by measuring their fertility on random samples and training splits, comparing them to two established German tokenizers: german-gpt2 (vocab size: \num{50266}) and gbert-large (vocab size: \num{31102}).
Fertility measures how many subwords represent a single original word, with a value of 1 indicating perfect segmentation \cite{rust-etal-2021-good,ali-etal-2024-tokenizer}.
Although differing vocabulary sizes complicate direct comparisons, the results still provide relative performance insights.

\begin{table}
        \centering
        \resizebox{0.75\linewidth}{!}{%
                \begin{tabular}{lrr}
                        \toprule
                        \textbf{Tokenizer} & \textbf{Head}  & \textbf{Middle} \\
                        \midrule
                        word count         & \num{46509357} & \num{80782685}  \\    \midrule
                        german-gpt2        & 1.68           & 1.72            \\
                        gbert-large        & 1.72           & 1.74            \\
                        \midrule

                        ours 1TB           & 2.27           & 2.27            \\
                        ours 2023-2021     & 2.07           & 2.10            \\
                        ours 2023\_14      & 1.76           & 1.80            \\
                        \bottomrule
                \end{tabular}
        }
        \caption{Fertility of our three tokenizers with different training data sizes in comparison to other German tokenizers on two unseen training data samples: one from head and one from middle partition.}
        \label{tab:tok}
\end{table}

\Cref{tab:tok} shows the fertility of all tokenizers on two unseen dataset snapshots.
Both german-gpt2 and gbert-large achieve the lowest fertility.
Notably, among our own tokenizers, the one trained on the smallest dataset (2023\_14) produces fewer tokens on average than those trained on larger datasets.
This suggests that a smaller dataset may enable more efficient tokenization by concentrating on the most frequent tokens, while larger datasets introduce greater variability and less efficient segmentation.
As seen in \Cref{tab:tok_common_freq}, the tokenizer trained on less data also appears to yield more meaningful subword segments.
Consequently, we selected the tokenizer trained on the 2023\_14 snapshot.
\cbox{Interesting}{Fitting a tokenizer on ``too much'' data can reduce its efficiency, possibly due to having to account for variation in the data.}
To validate this finding, we repeated the experiment using a variety of disjoint training datasets and different test sets, including snapshots from different time periods as well as random internet texts.
Interestingly, we consistently observed the same outcome across all variations.

\subsection{Pretraining Process}
During training, we regularly saved and evaluated checkpoints to monitor the training process (\cref{sec:checkpoint_eval}).
Intermediate checkpoints will be published to enable further analysis and comparison with other models \uponacceptance{}.

\subsubsection{LLäMmlein 120M}\label{sec:train_proc_120M}
\begin{table}
        \centering
        \resizebox{\linewidth}{!}{%
                \begin{tabular}{rccccccc}

                        \toprule
                        \textbf{Model} & \textbf{FactCl.}  & \textbf{EUParl}   & \textbf{PAWSX}    & \textbf{NLI}      & \textbf{DB Asp.}  & \textbf{WebCAGe}  \\
                        \midrule
                        \num{10000}    & 0.711             & 0.531             & 0.427             & 0.549             & 0.454             & 0.689             \\
                        \num{100000}   & 0.708             & 0.532             & 0.464             & 0.559             & 0.479             & 0.700             \\
                        \num{200000}   & 0.705             & 0.497             & 0.497             & 0.575             & 0.464             & 0.703             \\
                        \num{300000}   & 0.712             & 0.525             & 0.497             & 0.615             & 0.498             & 0.682             \\
                        \num{400000}   & 0.713             & 0.522             & 0.488             & 0.627             & 0.511             & \textbf{0.695}    \\
                        \num{466509}   & 0.711             & 0.538             & 0.489             & \textbf{0.629}    & \textbf{0.517}    & 0.687             \\
                        \midrule
                        german-gpt2    & 0.707             & 0.533             & 0.394             & 0.479             & 0.429             & 0.645             \\
                        gbert-base     & \textbf{0.751}    & \textbf{0.616}    & \textbf{0.561}    & 0.436             & \underline{0.478} & \underline{0.693} \\
                        bert-ger-cased & \underline{0.721} & \underline{0.607} & \underline{0.537} & \underline{0.490} & 0.480             & 0.679             \\

                        \bottomrule
                \end{tabular}
        }
        \caption{Results of LLäMmlein 120M checkpoints on six SuperGLEBer tasks compared to similarly sized german-gpt2, gbert-base and bert-base-german-cased}
        \label{tab:120M_progress}
\end{table}

We evaluated LLäMmlein 120M against german-gpt2, gbert-base, and bert-base-german-cased.
While it consistently outperformed the decoder-only german-gpt2 model, BERT-based models excelled in the first three tasks (FactClaiming, EuroParl, PAWSX), reflecting known limitations of autoregressive architectures in tasks like sequence tagging and sentence similarity \cite{pfister-hotho-2024-supergleber}.
However, LLäMmlein 120M demonstrated superiority in complex classification tasks, outperforming all models in NLI from checkpoint \num{10000} onward, with its best checkpoint exceeding bert-base-german-cased by 14\%.
It also closely matches the top scores for DB Aspect and WebCAGe classification.

Performance trends during pretraining varied by task.
We calculate the Spearman correlation coefficient $r$ to measure the strength and direction of the relationship between pretraining steps and task performance, and the corresponding $p$-value to assess the statistical significance of the correlation.
FactClaiming and EuroParl showed minimal variation, but PAWSX ($r$ = 0.607, $p$ = 0.04), NLI ($r$ = 0.947, $p$ < 0.0001), and DB Aspect ($r$ = 0.909, $p$ < 0.0001) displayed significant linear improvements.

Despite this, an Analysis of Variance (ANOVA) across all 29 SuperGLEBer benchmark tasks revealed no statistically significant performance improvements beyond the \num{300000} training checkpoint (\Cref{fig:progress_120m}).
In particular, average performance at checkpoints \num{300000} (0.693) and \num{466509} (0.699) only demonstrated small gains of 0.06 (\Cref{fig:progress_120m}), despite additional \num{166509} training steps ($\approx$349 billion tokens).
These findings problematize the marginal returns of extended training on downstream tasks, suggesting potential early convergence or benchmark limitations in capturing nuanced model improvements on average across tasks.
While LLäMmlein quickly reached a plateau for certain tasks, i.e.\ those that might require more basic structure recognition (FactClaiming/EuroParl), it continued to learn and improve on some more complex tasks.
Interestingly, contrary to the ``curse of monolingual models'' posited by \citet{finetasks}, which suggests monolingual models excel at LM but lack reasoning, our model demonstrates strong performance on deeper semantic tasks such as NLI and WebCAGe.
\cbox{Contradiction}{\citet{finetasks} suggests monolingual models excel at LM but often lack reasoning; ours appear strong in both.}

\subsubsection{LLäMmlein 1B}\label{sec:train_proc_1B}
\begin{table}
        \centering
        \resizebox{\linewidth}{!}{%
                \begin{tabular}{rccccccc}
                        \toprule
                        \textbf{Model}   & \textbf{FactCl.}  & \textbf{EUParl}   & \textbf{PAWSX}    & \textbf{NLI}      & \textbf{DB Asp.}  & \textbf{WebCAGe}  \\
                        \midrule
                        \num{10000}      & 0.735             & 0.708             & 0.461             & 0.642             & 0.563             & 0.677             \\
                        \num{100000}     & 0.734             & 0.662             & 0.511             & 0.709             & 0.607             & 0.699             \\

                        \num{500000}     & 0.733             & 0.712             & 0.539             & 0.734             & 0.613             & 0.720             \\
                        \num{1000000}    & \underline{0.750} & 0.697             & 0.540             & 0.740             & 0.629             & 0.756             \\
                        \num{1430512}    & 0.736             & \underline{0.713} & 0.526             & \underline{0.749} & 0.623             & \underline{0.765} \\
                        \midrule
                        Llama 3.2. 1B    & 0.665             & 0.537             & 0.551             & 0.603             & 0.557             & 0.689             \\
                        EuroLLM-1.7B     & 0.724             & 0.654             & 0.585             & 0.529             & 0.587             & 0.662             \\ \midrule
                        gbert-base       & \textbf{0.751}    & 0.616             & \underline{0.561} & 0.436             & 0.478             & 0.693             \\
                        mbart-large-50   & 0.723             & \textbf{0.727}    & 0.358             & 0.336             & 0.471             & 0.651             \\
                        gbert-large      & 0.747             & 0.636             & \textbf{0.654}    & 0.736             & 0.550             & 0.716             \\
                        leo-mistral-7b   & 0.741             & 0.649             & -                 & \textbf{0.807}    & \underline{0.664} & -                 \\
                        leo-hessianai-7b & 0.747             & -                 & -                 & -                 & \textbf{0.669}    & \textbf{0.781}    \\

                        \bottomrule
                \end{tabular}
        }
        \caption{Results of LLäMmlein 1B across multiple training checkpoints on six SuperGLEBer tasks, in comparison to the best-performing models and models with similar parameter size.
                Following SuperGLEBer, results of models that experienced out-of-memory (OOM) errors on an A100 80 GB are indicated with a ``-''.}
        \label{tab:1B_progress}
\end{table}
We compared LLäMmlein 1B's performance on the SuperGLEBer benchmark to the best-performing models for each task and similarly sized models (\Cref{tab:1B_progress}).
All models and checkpoints are evaluated after a task-specific finetuning, following SuperGLEBer evaluation protocol.
While not always securing the top spot, it remains competitive across tasks, even against much larger models.
As with the 120M model, LLäMmlein 1B trails in sentence similarity tasks like PAWSX. However, it achieves competitive results for EuroParl.
Examining task progress over time reveals noticeable improvements across all tasks, except for FactClaiming.
Compared to the LLäMmlein 120M model, Spearman correlation analysis indicated significant positive relationships between training time and performance for all remaining tasks. In particular, also for EuroParl ($r$ = 0.431, $p$ = 0.009) and WebCAGe ($r$ = 0.92, $p$ < 0.0001), suggesting that LLäMmlein 1B continues to benefit from extended training.
Nevertheless, across all SuperGLEBer tasks, the advantage of extended pretraining diminished after roughly 30\% of the pretraining data was processed.
From this state, despite a slow decline in loss, no significant improvements were observed across the 29 downstream tasks (\Cref{fig:progress_1b}).
To investigate further, we evaluated the checkpoint where SuperGLEBer performance plateaued, along with its instruction-tuned variants on generative tasks \cite{leo_translate}.
Interestingly, while SuperGLEBer performance stagnated, generative benchmark results (\Cref{tab:chat_versions_saturated}) continued to improve on average, likely due to enhanced autoregressive language modeling capabilities.
\cbox{Interesting}{While generative tasks benefit from further pretraining, other task types do no longer after about 30\% of the pretraining data.}

\subsection{Final Model Evaluation}
Detailed results for all SuperGLEBer tasks can be found in \Cref{tab:results}, and on the official website \url{https://lsx-uniwue.github.io/SuperGLEBer-site/leaderboard_v1}.

\subsubsection{LLäMmlein 120M}
After evaluating LLäMmlein's performance across pretraining, we compared its final results against other models on the full SuperGLEBer benchmark, including pairwise t-tests to compare results with other models on the SuperGLEBer benchmark.
As shown in \Cref{tab:120M_progress,fig:120M_gpt2}, the final checkpoint of LLäMmlein significantly outperforms german-gpt2, establishing itself as the leading German decoder model in this size range.
Against BERT-based models (gbert-base and bert-german-cased), no significant differences were found (\Cref{tab:120M_progress,fig:120M_gbert,fig:120M_bert}), despite BERT's known strengths in sequence tagging and similarity tasks \cite{pfister-hotho-2024-supergleber}.
This highlights LLäMmlein's ability to compete effectively with established BERT models, even with their architectural advantages.

We further evaluated our results on the lm-evaluation-harness-de evaluation benchmark for autoregressive models against german-gpt2, the only other German-only decoder model available at this parameter size (\Cref{tab:base_1b_leo}), and find that we outperform or closely match this model for all tasks, except for TruthfulQA.

\subsubsection{LLäMmlein 1B}
\begin{table}
        \centering
        \resizebox{\linewidth}{!}{%
                \begin{tabular}{lcccc}
                        \toprule
                        \textbf{Model}            & \textbf{Truth.QA
                        }                         & \textbf{ARC-Chal.}                       & \textbf{HellaSwag}                       &
                        \textbf{MMLU}                                                                                                                                                                                         \\
                        \midrule
                        german-gpt2               & 0.432                                    & 0.236                                    & 0.268                                    & 0.238                                    \\
                        ours 120M                 & 0.404                                    & 0.238                                    & 0.320                                    & 0.245                                    \\
                        \midrule

                        Llama 3.2 1B              & 0.407                                    & 0.310                                    & 0.412                                    & 0.284                                    \\
                        Llama 3.2 1B Inst.        & \textcolor[HTML]{4781b4}{\textbf{0.440}} & 0.296                                    & 0.411                                    & \textcolor[HTML]{4781b4}{\textbf{0.343}} \\
                        ours 1B                   & 0.365                                    & 0.311                                    & 0.483                                    & 0.253                                    \\
                        ours 1B Guanako           & 0.375                                    & 0.313                                    & \textcolor[HTML]{4781b4}{\textbf{0.502}} & 0.258                                    \\
                        ours 1B Alpaka            & 0.397                                    & \textcolor[HTML]{4781b4}{\textbf{0.323}} & 0.499                                    & 0.258                                    \\ \midrule

                        Llama 2 7b                & 0.422                                    & 0.381                                    & 0.513                                    & 0.400                                    \\
                        leo-hessianai-7b-chat     & 0.452                                    & 0.442                                    & 0.624                                    & 0.401                                    \\
                        Disco-Llama3-Ger-8B Inst. & \textbf{0.530}                           & \textbf{0.538}                           & \textbf{0.664}                           & \textbf{0.559}                           \\
                        em-german-7b-v01          & 0.427                                    & 0.233                                    & 0.276                                    & 0.241                                    \\
                        \bottomrule
                \end{tabular}
        }
        \caption{Performance of our (instruction tuned) models on the lm-evaluation-harness-de, with TruthfulQA (mc2), ARC-Challenge (acc\_norm), HellaSwag (acc\_norm), MMLU (acc).
                Short version of \Cref{tab:base_1b_leo_full}.}
        \label{tab:base_1b_leo}
\end{table}
We evaluated LLäMmlein 1B against similarly sized and larger models.
Compared to Llama 3.2 (1B) and EuroLLM (1.7B), LLäMmlein 1B consistently outperformed both (\Cref{fig:1B_comparison_llama321,fig:1B_comparison_euro}).
Leo-hessianai-7b , showed superior performance, reaffirming the size advantage of 7B models (\Cref{tab:base_1b_leo,fig:1B_comparison_leo}).
Interestingly, LLäMmlein 1B showed no significant difference in performance compared to other, larger models like the Disco-Llama3-German (8B), Llama 3.1 (8B), and gbert-large (\Cref{tab:base_1b_leo,fig:1B_comparison_llama38,fig:1B_comparison_llama318,fig:1B_comparison_gbert}), highlighting its efficiency and competitiveness.

\Cref{tab:base_1b_leo} compares LLäMmlein 1B and its instruction-tuned variants with Llama 3.2 1B and larger models.
Notably, the German-finetuned Disco-Llama 3 (8B) instruct model achieved the highest scores overall, showing the benefit of increased size and instruction tuning.
However, this model had no significant advantage over LLäMmlein 1B on SuperGLEBer, suggesting that on average model size matters more for generative tasks than for this benchmark.

For smaller models, Llama 3.2 (1B) achieved the best TruthfulQA score.
However, in completion tasks like ARC-Challenge and HellaSwag, LLäMmlein 1B  Instruct models consistently outperformed both the base LLäMmlein model and Llama 3.2 1B, indicating that instruct-tuning enhances performance on structured completion tasks.
Conversely, Llama 3.2 1B Instruct excelled in the broader knowledge-focused MMLU benchmark.
Interestingly, instruct-tuning improved LLäMmlein scores across all tasks, a trend not observed for the Llama 3.2 model.

Task-specific results highlighted structural differences.
While ARC-Challenge and HellaSwag focus on commonsense reasoning, TruthfulQA and MMLU emphasize factual understanding.
Smaller models, even when finetuned, struggle more with question-answering tasks.
Comparing the 120M and 1B versions, the latter consistently outperformed the smaller model by ~10\%, except for MMLU and TruthfulQA.
Interestingly, the 120M model outperforms the 1B model on TruthfulQA, aligning with \citet{lin-etal-2022-truthfulqa}, who found smaller models often beat their larger counterparts.
\cbox{Confirmation}{120M model is better at TruthfulQA than 1B model, confirming the findings of \citet{lin-etal-2022-truthfulqa}.}

We observed that scaling from 120M to 1B parameters yields only marginal improvements in sentence similarity and question answering tasks (GermanQuAD and MLQA), with performance differences below 2\% and 4\%, respectively (\Cref{tab:results}).
This contrasts with SuperGLEBer, where these tasks showed more significant scaling benefits.
\cbox{Contradiction}{Scaling provides fewer benefits for tasks like QA and sentence similarity, contradicting prior results from SuperGLEBer \cite{pfister-hotho-2024-supergleber}.}

\subsection{Checkpoint Averaging}\label{sec:result_averaging}
Checkpoint averaging did not improve -- or even change -- downstream task performance on SuperGLEBer for either the 120M or 1B model (see \Cref{fig:1b_average_chkp} for the 1B model).
This was unexpected, but we hypothesize the checkpoints being too far apart, as \citet{vaswani_attention_2017} averaged checkpoints written every 10 minutes near the end of training, while our checkpoints are about 6-8 hours apart.
\cbox{Interesting}{Checkpoint averaging ineffective, possibly checkpoints are too far apart.}

\section{Related Work}\label{sec:related_work}
\subsection{German LLMs and Their Limitations}
While several language models include German, relatively few have been trained exclusively on German data, and even fewer have transparently documented the process and model capabilities.

\paragraph{German-only Models} Early German-focused models were predominantly encoder-based (e.g., BERT variants) trained on corpora up to 163.4GB \cite{chan-etal-2020-germans}.
A GPT-2 style German model was trained on 16GB of mixed-domain data \cite{german-gpt2}.
Contemporaneous to our work, DOSMo~\cite{dosmo7b2024} introduced a Mistral-7B model trained on 1T tokens of German text from a variety of sources.
However, little details about DOSMo's training process, data filtering, and evaluation is publicly known.
Furthermore, after acceptance two ModernBERT models have been trained using our dataset \cite{moderngbert}.

\paragraph{Multi-/Crosslingual Models Including German}
Several multilingual models incorporate German data, including Büble~\cite{delobelle2024buble}, bloom-6b4-clp-german~\cite{Ostendorff2023clp}, GerPT2~\cite{Minixhofer_GerPT2_German_large_2020}, Disco-Llama3-German-8B \cite{DiscoResearch2024}, EuroLLM-1.7B~\cite{martins2024eurollmmultilinguallanguagemodels}, and leo-hessianai-7b~\cite{pluster2023leolm}, as well as mGPT~\cite{mGPT}, a multilingual variant of GPT-2.
While these models demonstrate the feasibility of German (transfer) language modeling, they typically offer limited transparency in German data preprocessing, training conditions, and systematic evaluation.
In contrast, our work is the first to
\begin{enumerate*}[label=(\arabic*)]
        \item train a German-only LLM fully from scratch,
        \item provide a detailed, transparent description of the training pipeline and data sources, and
        \item rigorously evaluate the resulting model's German capabilities.
\end{enumerate*}

\subsection{Comparable Efforts in Other Languages}
Transparent training and comprehensive evaluation have become more common in other language contexts.
Pythia~\cite{pythia}, for example, released a suite of English models with detailed training logs, and Latxa~\cite{etxaniz-etal-2024-latxa} continued pretraining Llama 2 models on Basque data (4.2B tokens), thus significantly improving the models' Basque language modelling capabilities, while openly documenting its setup and performance.
Furthermore, \citet{virtanen2019multilingual} show that explicitly pretraining models monolingually on Finnish is able to outperform multilingually trained models.
Our approach extends this ethos of openness and thorough evaluation to the German language, advancing both model quality and reproducibility.

\section{Conclusion}
We developed two German-only decoder models, LLäMmlein 120M and 1B, trained from scratch with tailored tokenization and preprocessing.
Throughout training, we evaluated intermediate checkpoints to analyze task-specific learning dynamics, noting varied speeds of improvement and early plateaus for some tasks.

On the SuperGLEBer benchmark, LLäMmlein 1B consistently matched or outperformed comparable models, including multilingual Llama 3.2 1B, highlighting the potential benefits of monolingual training for language-specific tasks.
While generative question answering revealed limitations of smaller models, our 1B model performed comparably to larger models in most tasks.

Future work includes deeper analysis of training dynamics using our published checkpoints and data, creating high-quality German instruct datasets, and exploring domain-specific fine-tuning for further improvement.

\section{Limitations}
While the LLäMmlein models represent a significant contribution to German NLP research, several limitations remain:
\begin{enumerate*}
        \item \textbf{Limited Capabilities on some domains}
              Due to the scarcity of high-quality German resources for e.g.\ coding, we found the models perform poorly on such tasks.
        \item \textbf{Monolingual Focus}
              While being considered a strength in the context of this setup, LLäMmlein lacks the ability to leverage multilingual contexts or perform cross-lingual tasks, which could limit usability in certain scenarios.
        \item \textbf{Evaluation Scope}
              While evaluated extensively on the SuperGLEBer benchmark and lm-evaluation-harness, other domains such as literature, spoken language, or dialects were not tested, leaving gaps in the understanding of model capabilities.
        \item \textbf{Long-Context Handling}
              The models were trained with a maximum sequence length of 2048 tokens, which limits their applicability to tasks requiring extended contexts, such as processing long documents or legal texts.
\end{enumerate*}

\section*{Acknowledgments}
The authors gratefully acknowledge the scientific support and HPC resources provided by the Erlangen National High Performance Computing Center (NHR@FAU) of the Friedrich Alexander-Universität Erlangen-Nürnberg (FAU) under the NHR project b185cb.
NHR funding is provided by federal and Bavarian state authorities.
NHR@FAU hardware is partially funded by the German Research Foundation (DFG) – 440719683.
Furthermore, we gratefully acknowledge the HPC resources provided by the JuliaV2 cluster at the Universität Würzburg (JMU), which was funded as DFG project as ``Forschungsgroßgerät nach Art 91b GG'' under INST 93/1145-1 FUGG.
The project staff is partially funded by the DFG -- 529659926.
The data science chair is part of the CAIDAS, the Center for Artificial Intelligence and Data Science, and is supported by the Bavarian High-Tech Agenda, which made this research possible.
We would also like to thank Anton Vlasjuk for his valuable contributions and insights, particularly for his efforts in integrating Flash Attention.

\bibliography{anthology_0,anthology_1,julia_ma,custom}

\appendix
\section{Exemplary Shortcoming of Current Multilingual Models in German}\label{app:exemp_shortcomings}

\begin{quote}
        \textbf{Llama 3.1 8B:}

        \textit{Was haben David Hasselhoff und Markus Söder gemein?} \\
        Die beiden haben in den 1980er Jahren die Fernsehserie "Miami Vice" gemeinsam gespielt. Hasselhoff spielte \textbf{Kraftedurch} die Figur von Ricardo Montalban, der \textbf{Cobrai} im Film "Zorba the Greek" gespielt hat; Söder war Lieutenant Vincent Lopiano und \textbf{wurde während des Drehprozesses an einen Verletzung erkrankt}.

        \textbf{Issues:}
        \begin{itemize}
                \item \textbf{Kraftedurch:} This name is nonsensical and seems like a verbatim translation of two English words.
                \item \textbf{Cobrai:} Like Kraftedurch a nonsensical name.
                \item \textbf{wurde während des Drehprozesses an einen Verletzung erkrankt:} Grammatically incorrect, mixing verb forms and incorrect article use.
        \end{itemize}
\end{quote}

\begin{quote}
        \textbf{Ours 1B chat\_selected:}

        \textit{Was haben David Hasselhoff und Markus Söder gemein?} \\
        Sowohl David Hasselhoff als auch Markus Söder sind zwei berühmte deutsche Persönlichkeiten, die sich in der Unterhaltungs- und Politikwelt einen Namen gemacht haben. Sie teilen einige gemeinsame Merkmale wie ihren Erfolg bei den Fans, ihre Popularität und ihren Sinn für Humor.
\end{quote}

Ignoring the possibly absurd content of this query, the responses highlight underlying issues with the German language of existing multilingual open-source LLMs.
The response from Llama 3.1 8B contains glaring grammatical errors and nonsensical phrases, such as ``Kraftedurch'' (a meaningless word) and ``wurde während des Drehprozesses an einen Verletzung erkrankt'' (grammatically incoherent), which highlight a lack of robust German language training.
These issues, which required minimal effort to uncover, exemplarily show the inadequacy of current large-scale LLMs for German\footnote{\url{https://www.reddit.com/r/LocalLLaMA/comments/1bfce18/still_didnt_found_a_better_small_german_llm_anyone/}}, where even basic grammatical rules are frequently violated.
This demonstrates the critical importance of dedicated, large-scale German LLM pretraining to address these shortcomings.

\section{Removed Datapoints\dots}\label{sec:appendix}
To improve the overall quality and diversity of the dataset, we applied additional paragraph-level deduplication, to remove repetitive and redundant boilerplate texts (\Cref{subsec:dedup}) and a token-to-word ratio filter (\Cref{subsec:filtering}), to further exclude low-quality content.
For instance, the following paragraphs were removed by our additional preprocessing steps:

\subsection{\dots from deduplication}\label{subsec:dedup}
\{ "raw\_content": ...Die Nutzung der im Rahmen des Impressums oder vergleichbarer Angaben veroeffentlichten Kontaktdaten wie Postanschriften, Telefon- und Faxnummern sowie Emailadressen durch Dritte zur Uebersendung von nicht ausdruecklich angeforderten Informationen ist nicht gestattet..., ...\}

\{ "raw\_content": ...5) Datenverarbeitung bei Eröffnung eines Kundenkontos Gemäß Art. 6 Abs. 1 lit. b DSGVO werden personenbezogene Daten im jeweils erforderlichen Umfang weiterhin erhoben und verarbeitet, wenn Sie uns diese bei der Eröffnung eines Kundenkontos mitteilen. Welche Daten für die Kontoeröffnung erforderlich sind, entnehmen Sie der Eingabemaske des entsprechenden Formulars auf unserer Website. ... \}

\subsection{\dots from tokenizer filtering}\label{subsec:filtering}
\{"raw\_content": "Home > B > Bamboo > Masaya 1Masaya 1 Guitar Tabs Masaya 1 Guitar Tabs Bamboo Do you like Masaya 1? Share with your friends now Bass TabsBass Tabs v2ChordsChords v2Chords v3Chords v4TabsTabs v2Ukulele Artist/band: Bamboo e|----3---3---3---3---3---3-3-3---0-0---0-0-0-0-0-0-0-0-0-0-0-0-0-0-0-------------| B|----3---3---3---3---3---3-3-3---0-0---1-1-1-3-3-3-3-p1-1-1-3-3-3-3-3------------|G|--------------------------0-0-0---0-0---2-2-2-2-2-2-2-2-2-2-2-2-2-2-------------|..., "doc\_id": "2014-52/0086/de\_head.json.gz/84", "quality\_signals": \{"ccnet\_perplexity": 41.2, ...\}\}
\newline
\newline
\{"raw\_content": DogsTootsie You are not logged in:
Owner: merrier\_with\_a\_terrierBreed: Wire Fox Terrier Gender: Female Jun 20, 2008twolfgirl66 Cute little girl!!!!!!!!!!!!!!!!!!!!!!! Jun 20, 2008 ttontosmommy aaaaaaaaaaaaaaaaaaaaaaaaaaaaaaaaaaa  aaaaaaaaaaaaaaaaaaaaaaaaaaaaaaaaaaaaaaaaaaa aaaaaaaaaaaaaaaaaaaaaaaaaaaaa
aaaaaaaaaaaaaaaaaaaaaaaaaaaaaaaaaaa aaaaaaaaaaaaaawwwwwwwwwwwwwww wwwwwwwwwwwwwwwwwww wwwwwwwwwwwwwwwwww wwwwwwwwwwwwww wwwwwwwwwwwwwwwwwwwwww...,
"doc\_id":2014-52/0058/de\_head.json.gz/309, ...\}

\section{Tokenizer}
To investigate the performance differences of our three trokenizer variants trained on different amounts of data, we analyzed the most frequently used tokens and the total number of unique subwords produced by each tokenizer on the head snapshot of 2014\_15 (\Cref{tab:tok_common_freq}).
As expected, all tokenizers shared common punctuation tokens (e.g., ``.'' and ``,'') among their most frequent entries.
However, notable distinctions emerged in how frequently used German words were tokenized.
The 2023\_14 tokenizer captures frequent German words like ``der'' and ``und'', whereas 2023-2021 and 1TB, exhibited a higher frequency of single-character tokens (e.g., ``e'', ``r'').
This pattern supports the hypothesis that the smaller dataset allows for a more efficient representation of frequently used tokens, while the larger datasets introduce more variability, leading to tokenization into smaller subunits.
Remarkably, the frequent tokens of 2023\_14 closely resembled those of german-gpt2 (vocab size (\num{50266}), reinforcing its alignment with established baselines in capturing essential German vocabulary.
Notably, the 2023\_14 tokenizer utilized nearly its entire vocabulary (\num{31959} out of \num{32000} tokens) when processing the unseen data, suggesting an effective distribution.

\begin{table}
        \centering
        \resizebox{\linewidth}{!}{%
                \begin{tabular}{lcl|cl|cl|cl}
                        \toprule
                                      & \multicolumn{2}{c}{\textbf{2023\_14}} & \multicolumn{2}{c}{\textbf{2023-2021}} & \multicolumn{2}{c}{\textbf{1TB}} & \multicolumn{2}{c}{\textbf{german-gpt2}}                                         \\
                        \textbf{Rank} & Token                                 & Frequency                              & Token                            & Frequency                                & Token & Frequency & Token & Frequency \\
                        \midrule
                        1.            & .                                     & 2.967.221                              & Ġ                                & 3.520.850                                & e     & 6.129.204 & .     & 2.964.871 \\
                        2.            & ,                                     & 2.535.194                              & .                                & 2.967.544                                & Ġd    & 4.474.530 & ,     & 2.538.127 \\
                        3.            & Ċ                                     & 1.941.957                              & e                                & 2.736.916                                & n     & 3.586.091 & Ċ     & 1.941.957 \\
                        4.            & Ġder                                  & 1.510.132                              & ,                                & 2.535.765                                & .     & 2.967.544 & Ġder  & 1.509.384 \\
                        5.            & Ġund                                  & 1.247.787                              & r                                & 2.256.061                                & i     & 2.926.354 & Ġund  & 1.247.544 \\
                        6.            & Ġdie                                  & 1.140.601                              & Ċ                                & 1.941.957                                & r     & 2.724.928 & Ġdie  & 1.140.601 \\
                        7.            & -                                     & 1.017.930                              & in                               & 1.808.081                                & ,     & 2.535.904 & -     & 1.022.213 \\
                        8.            & Ġin                                   & 826.190                                & Ġde                              & 1.547.525                                & Ġ     & 2.061.515 & Ġin   & 822.906   \\
                        9.            & Ġ(                                    & 600.305                                & nd                               & 1.281.432                                & Ċ     & 1.941.957 & Ġ(    & 599.928   \\
                        10.           & Ġvon                                  & 588.567                                & Ġu                               & 1.264.085                                & s     & 1.791.900 & Ġvon  & 588.567   \\
                        \midrule
                        unique        & \multicolumn{2}{c}{31.959}            & \multicolumn{2}{c}{31.568}             & \multicolumn{2}{c}{31.328}       & \multicolumn{2}{c}{49.723}                                                       \\
                        \bottomrule
                \end{tabular}%
        }
        \caption{Comparison of the most frequently used tokens on the ``2014\_52'' head snapshot. ``Unique'' gives the count of distinct tokens used to encode the unseen data.}
        \label{tab:tok_common_freq}
\end{table}

\section{Training}\label{app:training}
\begin{figure*}
        \centering
        \includegraphics[width=0.75\textwidth]{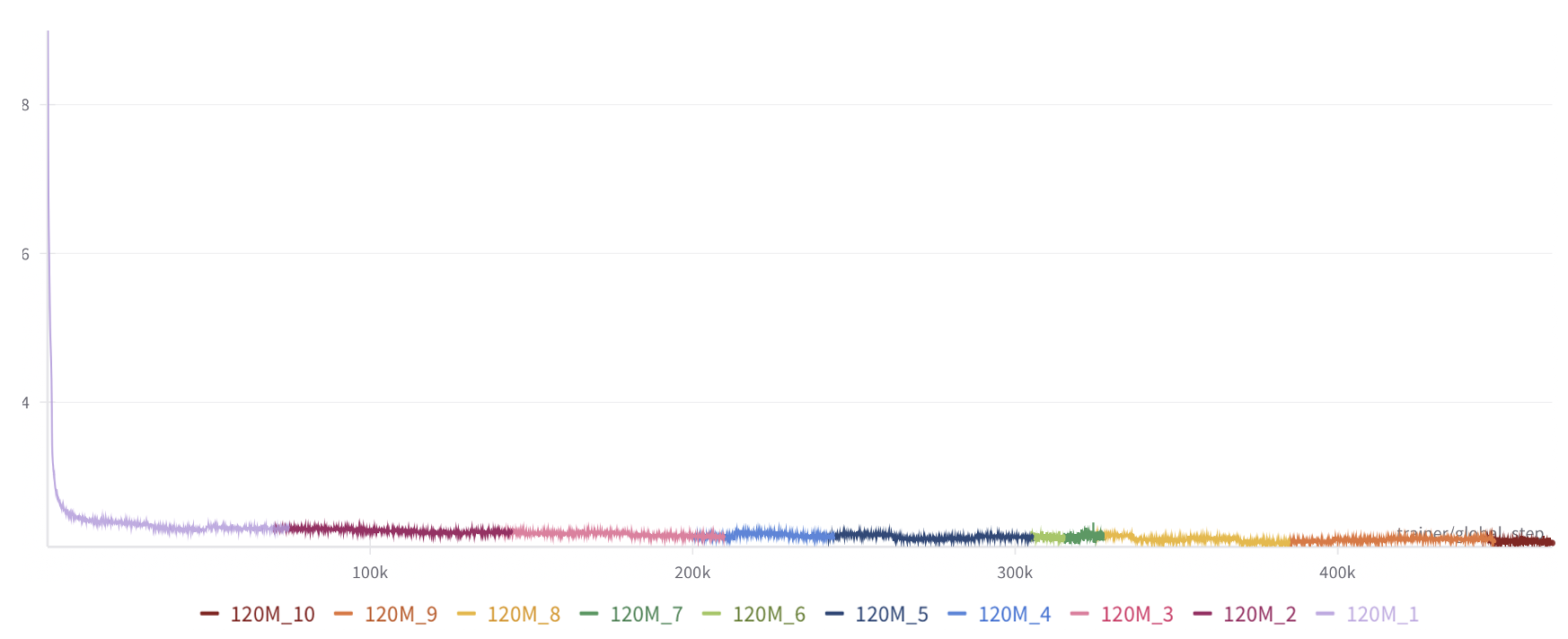}
        \caption{Loss curve of LLäMmlein 120M model.  Each color indicates a run, resumed after a training interruption.}\label{fig:loss_120M}
\end{figure*}

\begin{figure*}
        \centering
        \includegraphics[width=0.75\textwidth]{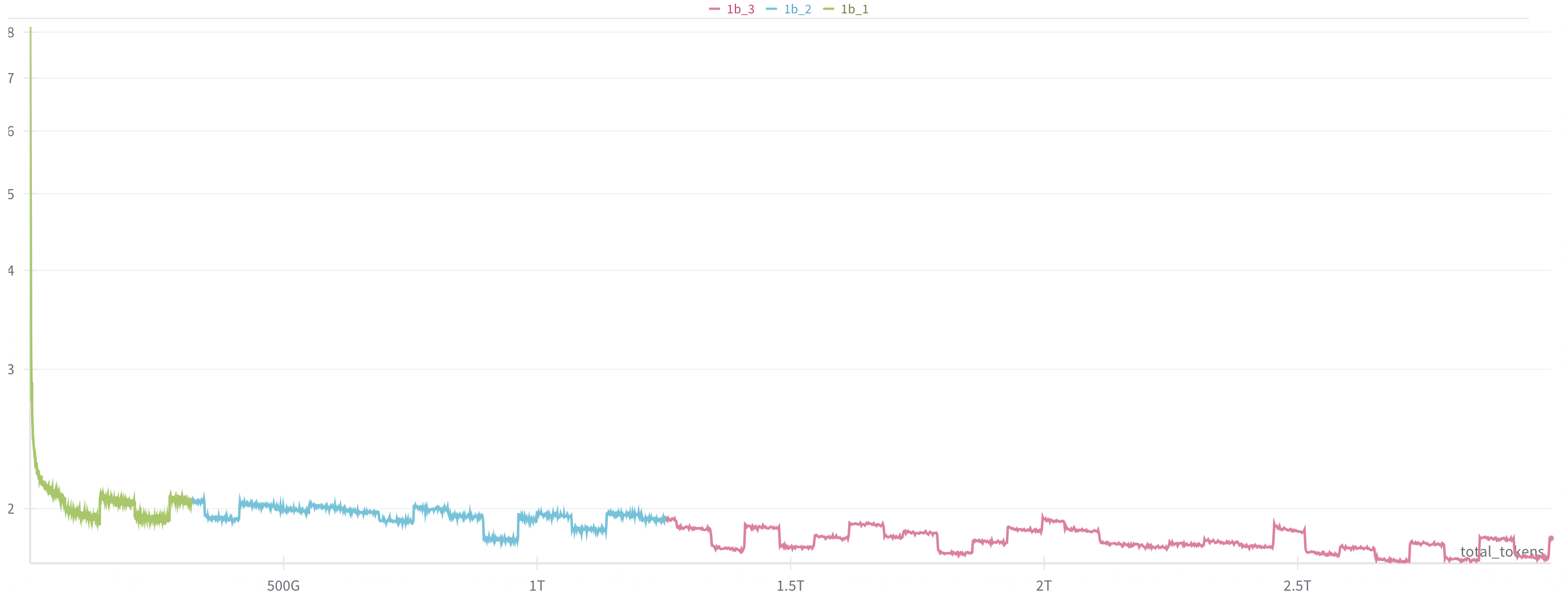}
        \caption{Loss curve of LLäMmlein 1B model. Each color indicates a run, resumed after an interruption. The visible jumps correspond to different chunks of training data, each sampled from a distinct part of the dataset.}
        \label{fig:loss_1b}
\end{figure*}

\begin{table}[h]
        \centering
        \resizebox{\linewidth}{!}{%
                \begin{tabular}{lcc}
                        \toprule
                                       & \textbf{LLäMmlein 120M} & \textbf{LLäMmlein 1B} \\
                        \midrule
                        Parameters     & \num{124668672}         & \num{1035638784}      \\
                        Heads          & 12                      & 32                    \\
                        Layer          & 12                      & 22                    \\
                        Tokens         & 1T                      & 3T                    \\
                        Training steps & \num{466509}            & \num{1430512}         \\
                        Learning rate  & 6e-4                    & 6e-4                  \\
                        Batch size     & 1024                    & 1024                  \\
                        Context length & 2048                    & 2048                  \\

                        \bottomrule
                \end{tabular}
        }
        \caption{Architectural and training details of LLäMmlein models}
        \label{tab:architectures}
\end{table}
Architectural and training details for both LLäMmlein models can be found in \Cref{tab:architectures}. In addition, we provide the loss curves for both models in \Cref{fig:loss_120M,fig:loss_1b}.
For LLäMmlein-120M overall, ten restarts were necessary:
Due to cluster settings, training was resumed at least every two days, and additionally, training had to be restarted a few times to address GPU and NCCL errors.
Before starting the training run for LLäMmlein-1B, we preliminary attempted to estimate the runtime for replicating the original TinyLlama training settings on our hardware as a sanity check.
Based on our extrapolations, the process would have taken over 200 days using 16 A100 GPUs, compared to the 90 days reported for TinyLlama on the same hardware.
After suspecting sharding configuration issues, we adapted the Fully Sharded Data Parallel (FSDP) strategy to a hybrid sharding approach.
This reduced the extrapolated runtime to approximately 100 days, which we deemed satisfactory.
Next, we scaled our training to the final 64 GPUs, bringing the extrapolated runtime down to 36 days.
Early in the training run, we identified further inefficiencies related to improper use of the available InfiniBand.
We halted the training, corrected the configuration, switched back to full sharding, and implemented dataset pre-caching in RAM on each node.
After these optimizations, we restarted the training, achieving a final overall runtime of 32 days (already including a few slower initial days before the final configuration change (green in \Cref{fig:loss_1b})), with a total of two restarts, illustrated in different colors in \Cref{fig:loss_1b}.
We will publish the code including all mentioned fixes/adaptations upon publication. 

\section{Evaluation on SuperGLEBer}
\begin{figure}
        \centering
        \includegraphics[width=\linewidth]{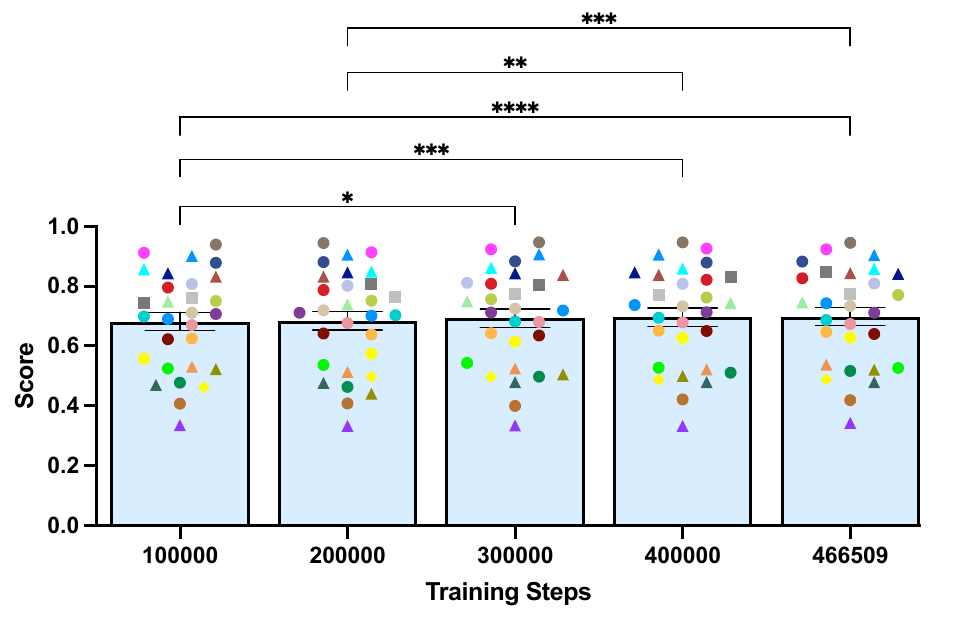}
        \caption{Statistical analysis of performance progress of 120M LLäMmlein over several checkpoints evaluated on the full SuperGLEBer dataset. Although all pairwise comparisons were calculate, non-significant connections (ns) were excluded for clarity.}
        \label{fig:progress_120m}
\end{figure}

\begin{figure}
        \centering
        \includegraphics[width=\linewidth]{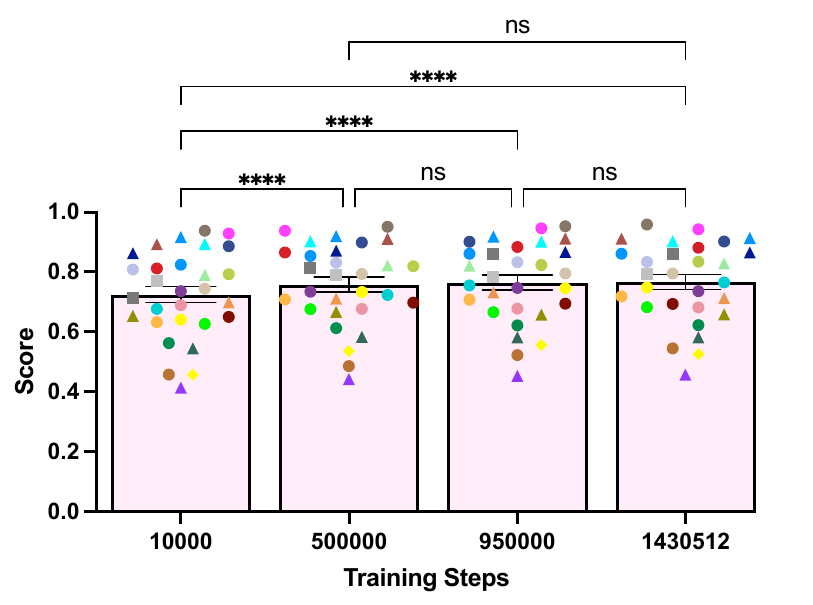}
        \caption{Statistical analysis of performance progress of 1B LLäMmlein over several checkpoints evaluated on the full SuperGLEBer dataset.}
        \label{fig:progress_1b}
\end{figure}

To investigate the influence of training steps on the model performance, we performed an Analysis of Variance (ANOVA) across multiple checkpoints that we evaluated on all 29 benchmark tasks.
For the 120M model no significant improvements were observable after the \num{300000} checkpoint (\Cref{fig:progress_120m}), while average performance plateaued after the \num{500000} checkpoint for the 1B model (\Cref{fig:progress_1b}), raising questions whether training could have been concluded earlier, or if further training still provides improvements uncaptured by the benchmark.

\Cref{tab:results} depicts concrete numbers for the SuperGLEBer benchmark comparing the reported encoder and decoder models with our final LLäMmlein 120M and 1B models, as well as their respective saturated variants.
LLäMmlein is competitive with models of the same parameter size and particularly excels in classification tasks, where the 1B model achieves the highest average score.
Notably, there is no significant difference observed between our models and its saturated counterparts.
Comparing the 120M and 1B model it is noticeable that the 1B LLäMmlein model shows clear superiority for classification and sequence tagging tasks compared to the 120M version.
However, this performance gap is smaller for question answering and sentence similarity tasks.
\input{result_table}

\subsection{120M vs.\ other models}
\begin{figure*}
        \centering
        \begin{subfigure}[t]{0.32\textwidth}
                \centering
                \includegraphics[width=\textwidth]{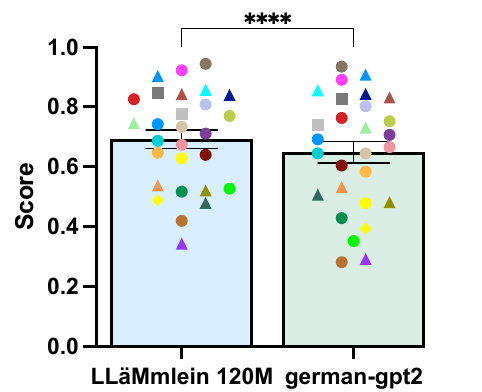}
                \caption{}\label{fig:120M_gpt2}
        \end{subfigure}
        \hfill
        \begin{subfigure}[t]{0.32\textwidth}
                \centering
                \includegraphics[width=\textwidth]{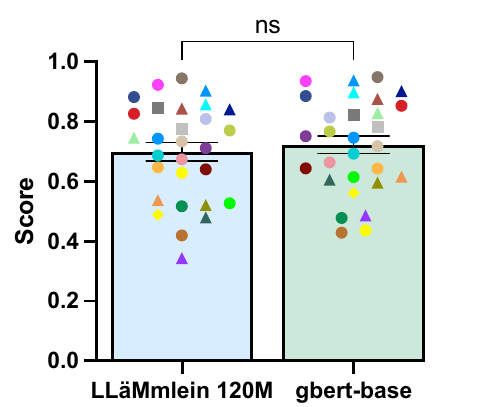}
                \caption{}\label{fig:120M_gbert}
        \end{subfigure}
        \hfill
        \begin{subfigure}[t]{0.32\textwidth}
                \centering
                \includegraphics[width=\textwidth]{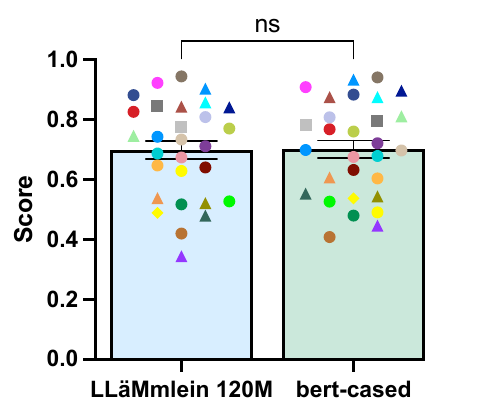}
                \caption{}\label{fig:120M_bert}
        \end{subfigure}
        \caption{Comparison of LLäMmlein 120M across the full SuperGLEBer benchmark with: (\ref{fig:120M_gpt2}) german-gpt2, (\ref{fig:120M_gbert}) gbert-base and (\ref{fig:120M_bert}) bert-base-german-cased.
                The asterisks indicate the level of statistical significance: ``ns'' denotes not significant (\(p > 0.05\)), while increasing significance is represented as follows: * (\(p \leq 0.05\)), ** (\(p \leq 0.01\)), *** (\(p \leq 0.001\)), and **** (\(p \leq 0.0001\)).}
        \label{fig:120M_comparison}
\end{figure*}

\Cref{fig:120M_comparison} illustrates comparisons (incl. t-tests) of our model against existing models of the same size on the SuperGLEBer benchmark.
LLäMmlein clearly outperforms german-gpt2, confirming its superiority among German decoder models of similar size.
When comparing LLäMmlein with the two BERT models -- gbert-base and bert-base-german-cased -- no statistically significant differences were found, which makes our 120M model the first to match the average performance of a similarly sized encoder on the SuperGLEBer benchmark.

\subsection{1B vs.\ other Models}
\begin{figure*}
        \centering
        \begin{subfigure}[t]{0.32\textwidth}
                \includegraphics[width=\textwidth]{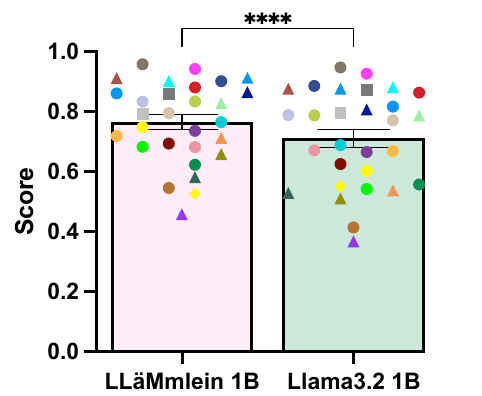}
                \caption{}\label{fig:1B_comparison_llama321}
        \end{subfigure}
        \begin{subfigure}[t]{0.32\textwidth}
                \includegraphics[width=\textwidth]{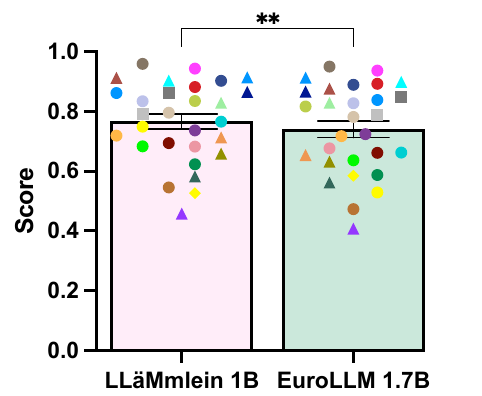}
                \caption{}\label{fig:1B_comparison_euro}
        \end{subfigure}
        \begin{subfigure}[t]{0.32\textwidth}
                \includegraphics[width=\textwidth]{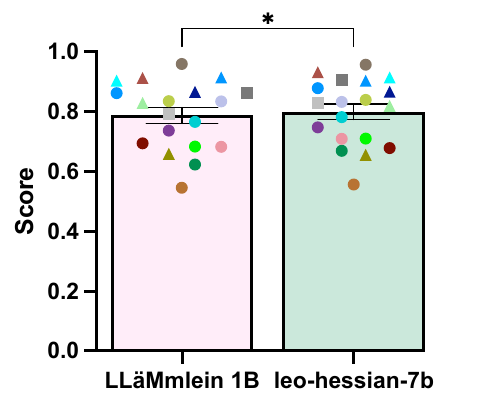}
                \caption{}\label{fig:1B_comparison_leo}
        \end{subfigure}

        \begin{subfigure}[t]{0.32\textwidth}
                \includegraphics[width=\textwidth]{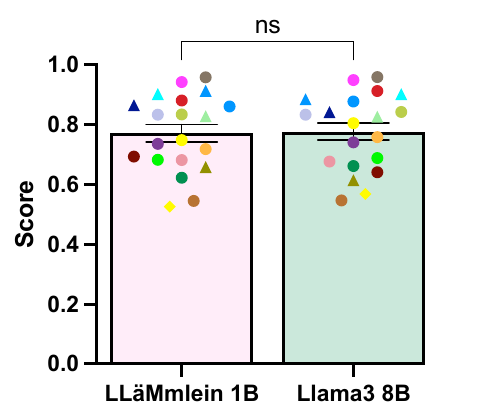}
                \caption{}\label{fig:1B_comparison_llama38}
        \end{subfigure}
        \begin{subfigure}[t]{0.32\textwidth}
                \includegraphics[width=\textwidth]{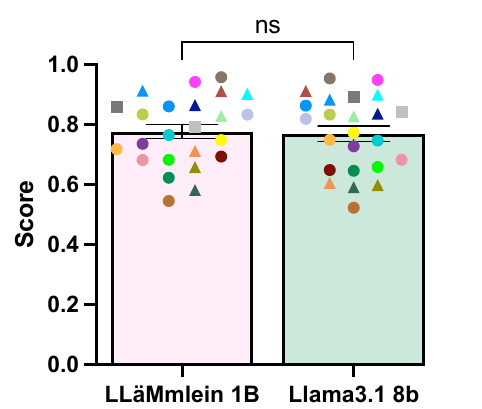}
                \caption{}\label{fig:1B_comparison_llama318}
        \end{subfigure}
        \begin{subfigure}[t]{0.32\textwidth}
                \includegraphics[width=\textwidth]{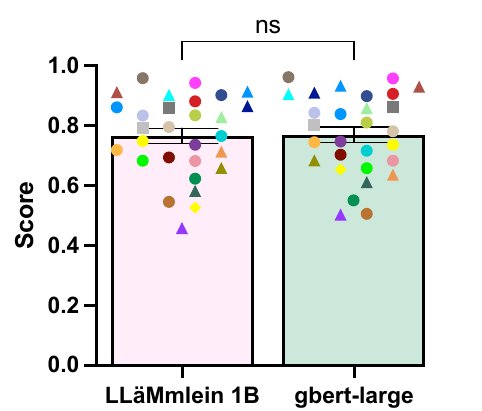}
                \caption{}\label{fig:1B_comparison_gbert}
        \end{subfigure}

        \caption{Performance comparison of LLäMmlein 1B accross the full SuperGLEBer benchmark with: (\ref{fig:1B_comparison_llama321}) Llama 3.2 1B, (\ref{fig:1B_comparison_euro}) EuroLLM-1.7B, (\ref{fig:1B_comparison_leo}) leo-hessianai-7b, (\ref{fig:1B_comparison_llama38}) on German-finetuned Disco-Llama 3 8B, (\ref{fig:1B_comparison_llama318}) Llama 3.1 8B and (\ref{fig:1B_comparison_gbert}) gbert-large.
                The asterisks indicate the level of statistical significance: ``ns'' denotes not significant (\(p > 0.05\)), while increasing significance is represented as follows: * (\(p \leq 0.05\)), ** (\(p \leq 0.01\)), *** (\(p \leq 0.001\)), and **** (\(p \leq 0.0001\)). For consistency, we entirely excluded tasks from the pairwise t-tests, where a larger model lacked a score due to a cuda out of memory error.}
        \label{fig:1B_comparison_1}
\end{figure*}

\Cref{fig:1B_comparison_1} compares our 1B model against models (using t-tests) of similar sizes and larger on the SuperGLEBer benchmark.
To ensure comparability, we excluded tasks in pairwise analysis, where one model lacked a score due to CUDA out-of-memory errors.
Among models with similar parameter sizes, we compare LLäMmlein 1B to Llama 3.2 with 1B parameters and EuroLLM with 1.7B parameters and significantly outperform them both.
While we were (expectedly) outperfomed by the seven times larger leo-hessianai-7b model, no significant performance differences were found between LLäMmlein 1B and other much larger models, such as the German-finetuned Diso-Llama 3 with 8B parameters, Llama3.1 8B and gbert-large.

\section{lm-evaluation-harness-de}
As decoder-only models are effective for generative tasks we further evaluated our models on the lm-evaluation-harness-de (\Cref{tab:base_1b_leo_full}).

\begin{table}
        \centering
        \resizebox{\linewidth}{!}{%
                \begin{tabular}{lcccc}
                        \toprule
                        \textbf{Model}            & \textbf{TruthfulQA
                        }                         & \textbf{ARC-Challenge}                         & \textbf{HellaSwag}                                                                &
                        \textbf{MMLU}                                                                                                                                                                                                                                                                                                                          \\
                        \midrule

                        german-gpt2               & 0.261|0.432                                    & 0.195|0.236                                                                       & 0.262|0.268                                                                       & 0.238|0.263                                                                       \\

                        ours 120M                 & 0.247|0.404                                    & 0.194|0.238                                                                       & 0.291|0.320                                                                       & 0.245|0.276                                                                       \\
                        \midrule
                        Llama 3.2 1B              & \textcolor[HTML]{4781b4}{\textbf{0.280}}|0.407 & 0.265|0.310                                                                       & 0.339|0.412                                                                       & 0.284|0.302                                                                       \\
                        Llama 3.2 1B Instruct     & 0.279|\textcolor[HTML]{4781b4}{\textbf{0.440}} & 0.259|0.296                                                                       & 0.340|0.411                                                                       & \textcolor[HTML]{4781b4}{\textbf{0.343}}|\textcolor[HTML]{4781b4}{\textbf{0.343}} \\

                        ours 1B                   & 0.239|0.365                                    & 0.266|0.311                                                                       & 0.390|0.483                                                                       & 0.253|0.270                                                                       \\
                        ours 1B full              & 0.257|0.388                                    & 0.282|0.318                                                                       & 0.395|0.499                                                                       & 0.254|0.273                                                                       \\
                        ours 1B Alpaka            & 0.268|0.397                                    & 0.279|0.323                                                                       & 0.399|0.499                                                                       & 0.258|0.273                                                                       \\
                        ours 1B Evol              & 0.255|0.378                                    & \textcolor[HTML]{4781b4}{\textbf{0.284}}|\textcolor[HTML]{4781b4}{\textbf{0.323}} & 0.397|0.498                                                                       & 0.250|0.268                                                                       \\
                        ours 1B Guanako           & 0.257|0.385                                    & 0.280|0.314                                                                       & 0.394|0.498                                                                       & 0.260|0.275                                                                       \\
                        ours 1B Sharegpt          & 0.242|0.371                                    & 0.275|0.317                                                                       & \textcolor[HTML]{4781b4}{\textbf{0.398}}|\textcolor[HTML]{4781b4}{\textbf{0.504}} & 0.250|0.270                                                                       \\

                        \midrule

                        Llama 2 7b                & 0.268|0.422                                    & 0.333|0.381                                                                       & 0.396|0.513                                                                       & 0.400|0.396                                                                       \\
                        leo-hessianai-7b-chat     & 0.301|0.452                                    & 0.405|0.442                                                                       & 0.485|0.624                                                                       & 0.401|0.401                                                                       \\
                        Disco-Llama3-Ger-8B       & 0.331|0.495                                    & 0.456|0.497                                                                       & 0.491|0.654                                                                       & 0.545|0.529                                                                       \\
                        Disco-Llama3-Ger-8B Inst. & \textbf{0.364}|\textbf{0.530}                  & \textbf{0.506}|\textbf{0.538}                                                     & \textbf{0.515}|\textbf{0.664}                                                     & \textbf{0.559}|\textbf{0.555}                                                     \\
                        em-german-7b-v01          & 0.225|0.427                                    & 0.197|0.233                                                                       & 0.258|0.276                                                                       & 0.241|0.263                                                                       \\
                        \bottomrule
                \end{tabular}
        }
        \caption{Performance comparison of (instruction tuned) LLäMmlein variants as well as similar sized and various larger models on the lm-evaluation-harness-de including the: TruthfulQA (mc1|mc2), ARC-Challenge (acc|acc\_norm), HellaSwag (acc|acc\_norm) and MMLU (acc|acc\_norm). This is the full table of \Cref{tab:base_1b_leo}, including all metrics and all trained instruct adapters.}
        \label{tab:base_1b_leo_full}
\end{table}

During analysis of the training process, we found no siginificant average performance improvement on the SuperGLEBer benchmark for the 1B model starting from the \num{500000} checkpoint.
To further investigate, we evaluated this checkpoint on the lm-eval-harness as well (\Cref{tab:chat_versions_saturated,sec:train_proc_1B}).

\begin{table}
        \centering
        \resizebox{\linewidth}{!}{%
                \begin{tabular}{lcccc}
                        \toprule
                        \textbf{Model}         & \textbf{TruthfulQA
                        }                      & \textbf{ARC-Challenge} & \textbf{HellaSwag} &
                        \textbf{MMLU}                                                                                    \\
                        \midrule
                        ours 1B\_sat. full     & 0.256|0.495            & 0.205|0.247        & 0.252|0.258 & 0.227|0.250 \\
                        ours 1B\_sat. Alpaka   & 0.273|0.494            & 0.213|0.255        & 0.253|0.259 & 0.229|0.251 \\
                        ours 1B\_sat. Evol     & 0.261|0.501            & 0.211|0.249        & 0.254|0.256 & 0.229|0.253 \\
                        ours 1B\_sat. Guanako  & 0.264|0.501            & 0.224|0.246        & 0.251|0.261 & 0.231|0.255 \\
                        ours 1B\_sat. Sharegpt & 0.262|0.495            & 0.202|0.243        & 0.255|0.261 & 0.230|0.249 \\

                        \bottomrule
                \end{tabular}
        }
        \caption{Performance comparison of ``saturated'' LLäMmlein 1B instruction tuned variants from checkpoint \num{500000}, from which no significant improvement was noticeable on the SuperGLEBer benchmark (has to be compared to \Cref{tab:base_1b_leo_full}).}
        \label{tab:chat_versions_saturated}
\end{table}

A finding of the translated lm-evaluation-harness-de:
During testing, we identified several instances of residual English text.
For example, in the "high\_school\_computer\_science" section:

\{'question': 'In Python 3, which of the following function convert a string to an int in python?', 'choices': ['int(x [,base])', 'long(x [,base] )', 'float(x)', 'str(x)'], 'answer': 0, 'question\_de': 'In Python 3, which of the following function convert a string to an int in python?', 'choices\_de': ['int(x [,base])', 'long(x [,base] )', 'float(x)', 'str(x)'], ...\}

and in the "machine\_learning" section:

\{'question': '\_ refers to a model that can neither model the training data nor generalize to new data.', 'choices': ['good fitting', 'overfitting', 'underfitting', 'all of the above'], 'answer': 2, 'question\_de': '\_\_\_\_\_\_\_ refers to a model that can neither model the training data nor generalize to new data.', 'choices\_de': ['gute Anpassung', 'Überanpassung', 'Unteranpassung', 'alle oben genannten'], ...\}'\}

Thus, the results and their information value about German language capabilities may not be completely accurate.

\begin{table*}[h!]
        \centering
        \label{tab:random_sample}
        \resizebox{\textwidth}{!}{%
                \begin{tabular}{@{}lll@{}}
                        \toprule
                        Prompt & Variant  & Output                                                                                                                                                                       \\ \midrule
                        \multirow{3}{*}{In Berlin\dots}
                               & German   & In Berlin-Mitte ist ein Mann mit einem Messer angegriffen und schwer verletzt worden.                                                                                        \\
                               & Bavarian & In Berlin is a Stod in da Bundesstod Berlin.                                                                                                                                 \\
                               & Swiss    & In Berlin ist es heute Nacht zu einem Brand gekommen.                                                                                                                        \\ \midrule
                        \multirow{3}{*}{In Minga\dots}
                               & German   & In Minga, der Stadt der Münchner, gibt es viele Sehenswürdigkeiten, die man sich anschauen kann.                                                                             \\
                               & Bavarian & In Minga is a Gmoa in Bayern.                                                                                                                                                \\
                               & Swiss    & In Minga Minga ([\textipa{mI\ng ga}]) isch en Begriff, wo im südliche Afrika, in de südliche Sahara und in de südliche Halbinse, wo in de Sahelzone liegt, gebräuchlig isch. \\ \midrule
                        \multirow{3}{*}{In Züri\dots}
                               & German   & In Züri West, dem neuen Album von Züri West.                                                                                                                                 \\
                               & Bavarian & In Züri is a Stod in der Schweiz.                                                                                                                                            \\
                               & Swiss    & In Züri isch es nid so heiss wie in Basel.                                                                                                                                   \\ \bottomrule
                \end{tabular}}
        \caption{Example outputs in German, Bavarian, and Swiss dialects for the base model and two adapter variants, when being asked about the ``regional captial'', showcasing linguistic adaptability across regional varieties. The outputs demonstrate varying levels of accuracy in adapting to the target dialects. While some results capture the linguistic characteristics effectively, others fall short, highlighting the challenges of fine-tuning models for nuanced regional variations. The tendency of the Bavarian adapter to adhere to the Wikipedia style is pronounced as it starts with a short descriptive sentence every time.}
\end{table*}

\section{Checkpoint Averaging}
\begin{figure}
        \centering
        \includegraphics[width=0.9\linewidth]{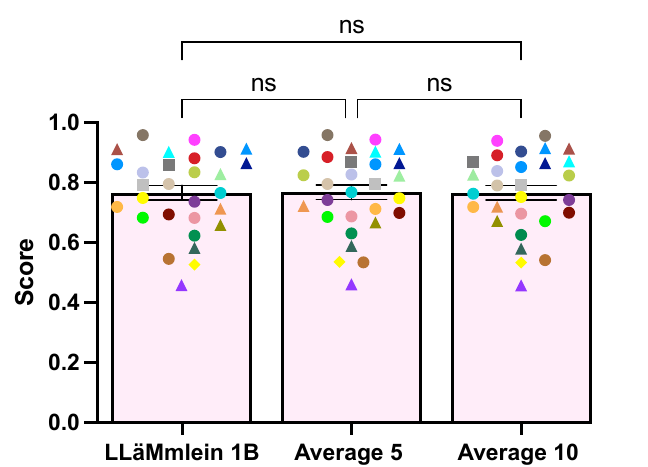}
        \caption{Comparison of results for the LLäMmlein 1B model and its five and ten checkpoints-averaged versions.}\label{fig:1b_average_chkp}
\end{figure}

Aiming to boost our performance, we experimented with checkpoint averaging (\Cref{sec:result_averaging}).
However, no significant differences were observable when comparing the final 1B LLäMmlein model with averaged checkpoints from the last five or ten checkpoints on the SuperGLEBer benchmark (\Cref{fig:1b_average_chkp}).
A possible reason for this could be, that the checkpoints were saved with too large intervals between them, as we recorded approximately three checkpoints per day.
In contrast, in the transformer paper \cite{vaswani_attention_2017}, where this technique was successfully used, checkpoints were saved every ten minutes.

\section{Exemplarily Dialectic Analysis}\label{app:dialectic_exploration}
To showcase the model's versatility, we finetuned adapters on specific dialectal data.
For Bavarian, we used approximately \num{25700} Wikipedia pages from the Bavarian column of the ``cis-lmu/bavarian\_to\_english'' dataset on HuggingFace.
For Swiss German, we finetuned an adapter on \num{206047} documents from the ``HuggingFaceFW/fineweb-2'' dataset.
Following the setup in \Cref{sec:exp_downstream}, these experiments, though not quantitatively evaluated, highlight the model's adaptability and potential, as illustrated in \Cref{tab:random_sample}.

\end{document}

%% file: result_table.tex
\begin{sidewaystable*}
    \centering
    \begin{tabular}{rrr|ccccc|c||cc|c||c||c}
        \toprule
                                                            &                                                                &                &
        \multicolumn{6}{c||}{classification}                &
        \multicolumn{3}{c||}{tagging}                       & \multirow{4}{*}{\rotatebox[origin=c]{270}{\makecell{similarity                                                                                      \\pearson corr}}}                               & \multirow{4}{*}{\rotatebox[origin=c]{270}{\makecell{QA                                                                                                                                                                       \\m.\ t.\ F1}}}                                                                                                                                                  \\
        type                                                & model                                                          &                &
        \rotatebox[origin=c]{270}{\makecell{tox.                                                                                                                                                                  \\macro F1}}     &
        \rotatebox[origin=c]{270}{\makecell{sent.                                                                                                                                                                 \\micro F1}}            &
        \rotatebox[origin=c]{270}{\makecell{match                                                                                                                                                                 \\ACC}}                 &
        \rotatebox[origin=c]{270}{\makecell{WSD                                                                                                                                                                   \\micro F1}}              &
        \rotatebox[origin=c]{270}{\makecell{other                                                                                                                                                                 \\mixed}}                     &
        \rotatebox[origin=c]{270}{\makecell{avg                                                                                                                                                                   \\mixed}}                       &
        \rotatebox[origin=c]{270}{\makecell{NER                                                                                                                                                                   \\micro F1}}                       &
        \rotatebox[origin=c]{270}{\makecell{other                                                                                                                                                                 \\micro F1}}                     &
        \rotatebox[origin=c]{270}{\makecell{avg                                                                                                                                                                   \\micro F1}}                       &
        \\ \midrule[.75pt]
        \multirow{4}{*}{\rotatebox[origin=c]{270}{encoder}} &
        gbert-base                                          &
                                                            & 0.537                                                          & 0.620          & 0.738          & 0.814          & 0.749          & 0.723
                                                            & 0.705                                                          & 0.806          & 0.786
                                                            & 0.561                                                          & 0.803                                                                              \\
                                                            &
        gbert-large                                         &
                                                            & 0.604                                                          & 0.673          & 0.811          & 0.837          & 0.816          & 0.785
                                                            & 0.744                                                          & 0.813          & 0.799
                                                            & 0.654                                                          & 0.833                                                                              \\
                                                            &
        gottbert                                            &
                                                            & 0.553                                                          & 0.607          & 0.753          & 0.806          & 0.609          & 0.635$\dagger$
                                                            & 0.666$\dagger$                                                 & 0.794          & 0.768
                                                            & 0.553                                                          & 0.795                                                                              \\
                                                            &
        bert-base-german-cased                              &
                                                            & 0.520                                                          & 0.589          & 0.690          & 0.794          & 0.745$\dagger$ & 0.710$\dagger$
                                                            & 0.678                                                          & 0.788          & 0.766
                                                            & 0.537                                                          & 0.789                                                                              \\ \cmidrule{2-14}

        \multirow{3}{*}{\rotatebox[origin=c]{270}{decoder}} &
        german-gpt2                                         &
                                                            & 0.443                                                          & 0.511          & 0.664          & 0.768          & 0.612          & 0.606
                                                            & 0.617                                                          & 0.725          & 0.704
                                                            & 0.394                                                          & 0.784                                                                              \\
                                                            &
        bloomz-560m                                         &
                                                            & 0.440                                                          & 0.472          & 0.748          & 0.730          & 0.562$\dagger$ & 0.575$\dagger$
                                                            & 0.388                                                          & 0.638          & 0.588
                                                            & 0.459                                                          & 0.788                                                                              \\
                                                            &
        leo-hessianai-7b                                    &
                                                            & 0.617                                                          & 0.739          & 0.000$\dagger$ & 0.391$\dagger$ & 0.589$\dagger$ & 0.534$\dagger$
                                                            & 0.481$\dagger$                                                 & 0.671$\dagger$ & 0.633$\dagger$
                                                            & 0.000$\dagger$                                                 & 0.867                                                                              \\ \cmidrule{2-14}
        \multirow{4}{*}{\rotatebox[origin=c]{270}{ours}}    &
        LLäMmlein 120M                                      &
                                                            & 0.530                                                          & 0.605          & 0.736          & 0.805          & 0.770          & 0.734
                                                            & 0.626                                                          & 0.737          & 0.714
                                                            & 0.489                                                          & 0.811                                                                              \\
                                                            &
        saturated 120M                                      &
                                                            & 0.518                                                          & 0.599          & 0.725          & 0.803          & 0.767          & 0.728
                                                            & 0.620                                                          & 0.736          & 0.713
                                                            & 0.497                                                          & 0.790                                                                              
                                                                  \\
                                                            &
        LLäMmlein 1B                                        &
                                                            & 0.619                                                          & 0.714          & 0.798          & 0.854          & 0.818          & 0.792
                                                            & 0.739                                                          & 0.785          & 0.776
                                                            & 0.526                                                          & 0.826                                                                              \\ &
        saturated 1B                                        &
                                                            & 0.592                                                          & 0.703          & 0.789          & 0.831          & 0.812          & 0.781
                                                            & 0.739                                                          & 0.784          & 0.775
                                                            & 0.536                                                          & 0.802                                                                            
                      \\
        \bottomrule

    \end{tabular}
    \caption{%
        SuperGLEBer results, averaged at varying levels of granularity, following \cite{pfister-hotho-2024-supergleber}.
        The columns reading ``avg'' have been averaged across the averages of the respective task types, in order to not overweight any task type for which more datasets exist, i.e.\ all ``NER'' tasks have been averaged into a single value before averaging across all tagging tasks.
        The second row gives the type of metric used for the respective task type.
        Here ``mixed'' means that at least two kind of metrics have been averaged together.
        The results marked with $\dagger$ have been averaged over tasks for which a ``Cuda OOM'' error occured on an A100 80GB GPU as well at all averages this affects transitively.
        All missing values have been treated as a 0.0 when calculating the average.
        ``saturated'' indicates our models at a checkpoint from which no more significant improvement on SuperGLEBer could be measured (\Cref{sec:train_proc_120M,sec:train_proc_1B,fig:progress_120m,fig:progress_1b}.
    }\label{tab:results}
\end{sidewaystable*}